# A MACHINE LEARNING BASED ANALYTICAL FRAMEWORK FOR SEMANTIC ANNOTATION REQUIREMENTS

Hamed Hassanzadeh[1] and MohammadReza Keyvanpour[2]

[1]Department of Electronic, Computer and IT, Islamic Azad University, Qazvin Branch, Qazvin, Iran
and member of Young Researchers Club
h.hassanzadeh@qiau.ac.ir

[2]Department of Computer Engineering, Alzahra University, Tehran, Iran
keyvanpour@alzahra.ac.ir

## ABSTRACT

*The Semantic Web is an extension of the current web in which information is given well-defined meaning. The perspective of Semantic Web is to promote the quality and intelligence of the current web by changing its contents into machine understandable form. Therefore, semantic level information is one of the cornerstones of the Semantic Web. The process of adding semantic metadata to web resources is called Semantic Annotation. There are many obstacles against the Semantic Annotation, such as multilinguality, scalability, and issues which are related to diversity and inconsistency in content of different web pages. Due to the wide range of domains and the dynamic environments that the Semantic Annotation systems must be performed on, the problem of automating annotation process is one of the significant challenges in this domain. To overcome this problem, different machine learning approaches such as supervised learning, unsupervised learning and more recent ones like, semi-supervised learning and active learning have been utilized. In this paper we present an inclusive layered classification of Semantic Annotation challenges and discuss the most important issues in this field. Also, we review and analyze machine learning applications for solving semantic annotation problems. For this goal, the article tries to closely study and categorize related researches for better understanding and to reach a framework that can map machine learning techniques into the Semantic Annotation challenges and requirements.*

## 1. INTRODUCTION

The majority of today's World Wide Web's content is designed for humans to read and understand, not for machines and computer programs to manipulate meaningfully. Computers can adeptly parse Web pages for layout and routine processing but, in general, machines have no reliable way to process the semantics. In addition, the number of web pages is increasing dramatically each day so the keyword based search engines cannot help users to find out their interest in an efficient way. The Semantic Web is an extension of the World Wide Web, in which information is given well-defined meaning, better enabling computers and people to work in cooperation [1]. The idea of semantic web is to leave most of tasks and decisions to machines. This is applicable with adding knowledge to web contents by understandable languages for machine and establish intelligent software agents that able to process this information. On the other hand, while the Semantic Web consists of structured information and explicit metadata, it paves the way to rapidly access information and ability of semantic search.

In a semantic based environment, to ensure that all the machines have a common understanding from metadata tags and to be able to communicate and cooperate to each other, there is a need for a shared repository that defines all the concepts. In semantic Web, ontology acts as this shared repository of semantics [2]. An ontology is commonly defined as an explicit, formal specification of a shared conceptualization of a domain of interest. This means that an ontology describes some application-relevant part of the world in a machine-understandable way [3]. In other words, ontology is considered as a tool that defines additional meanings that tagged to web pages and makes them available to be used by software agents and web applications [4].

The Semantic Web vision is of a Web in which resources are accessible not only to humans, but also to automated processes. The automation of tasks depends on elevating the status of the web from machine-readable to something we might call machine-understandable. The key idea is to have data on the web defined and linked in such a way that its meaning is explicitly interpretable by software processes rather than just being implicitly interpretable by humans.

To realize this vision, it will be necessary to associate metadata with web resources. One mechanism for associating such metadata is annotation. In particular, we may wish to annotate resources with semantic metadata that provides some indication of the content of a resource. This is a further step along the way from simple textual annotations, as the intention within the Semantic Web context is that this information will be accessible not only to humans but also to software agents [5]. The process of adding these metadata is called Semantic Annotation.

Regarding to large amount of documents that must be annotated in a wide spread domain such as the Web, it's obvious that manually annotating is would be an expensive, time consuming, and generally inefficient task. So, one of the most serious problems in semantic annotation domain is to automate this process. One way to handle this problem is to utilize machine learning techniques.

Machine Learning is the study of computer algorithms that improve automatically through experiences [6]. Various learning techniques are classified in for groups, i.e. supervised learning, unsupervised learning, semi-supervised learning, and active learning. Different machine learning approaches have been proposed for semantic annotation automation [7, 8].

In this paper, at first we present an inclusive layered classification of divers semantic annotation challenges and demonstrate that automation is one of the most important issues in this field. Then, we introduce an analytical framework which collects and closely study the approaches that use different machine learning techniques. This framework can give a guideline for future researches on the Semantic Web.

The rest of this paper is organized as follow; section 2 reviews the semantic annotation problem and its tasks and goals. Section 3 briefly reviews some related works. In section 4 a classification of semantic annotation challenges is presented. In section 5 we present and discuss our analytical framework. And section 6 presents our conclusions and directions for future works.

## 2. SEMANTIC ANNOTATION

In general, the annotation defines as the process of adding notes and comments to documents, images, or any resources. In the Web domain, annotation means adding information such as notes, commentary, links to source material, and so on, to existing web-accessible documents without changing the originals [9]. These annotations are meant to be shareable, also over the network, although notes would be useful even if they couldn't be shared. The annotation process can be done manually, automatically, and semi-automatically. Concisely, semantic annotation means appending machine understandable metadata to resources. We consider Semantic Annotation the idea of assigning to the entities in the text links to their semantic descriptions [10]. Manual annotation is more easily accomplished today, using authoring tools such as Semantic Word [11], which provide an integrated environment for simultaneously authoring and annotating text. However, the use of human annotators is often fraught with errors due to factors such as annotator familiarity with the domain, amount of training, personal motivation and complex schemas. Manual annotation is also an expensive process, and often does not consider that multiple perspectives of a data source, requiring multiple ontologies, can be beneficial to support the needs of different users [12]. By considering the large number of web documents and wide range of domains, it is obvious that semantic annotation task and beside it ontology development and enhancement, cannot be done in a manual and concentrative manner. Generally, the ineffectiveness of manual annotation can be described in these two conditions:

- It's cumbersome and time consuming; because of large amount of tasks and resources,

- It's objective; different opinions can result in inconsistent knowledge.

Semantic annotation in a manual manner can easily result in a knowledge enhancement bottleneck [13]. For facing this problem different automatic and semi-automatic approaches are introduced. In Figure 1 an overview of semantic annotation and the effective technologies in it is depicted.

Ontologies are the key elements of the most semantic annotation systems. Ontological structures may give additional value to semantic annotations. They allow for additional possibilities on the resulting semantic annotations, such as inferencing or conceptual navigation that we have mentioned before. But also the reference to a commonly agreed set of concepts by itself constitutes an additional value through its normative function. Furthermore, an ontology directs the attention of the annotator to a predefined choice of semantic structures and, hence, gives some guidance about what and how items residing in the documents may be annotated.

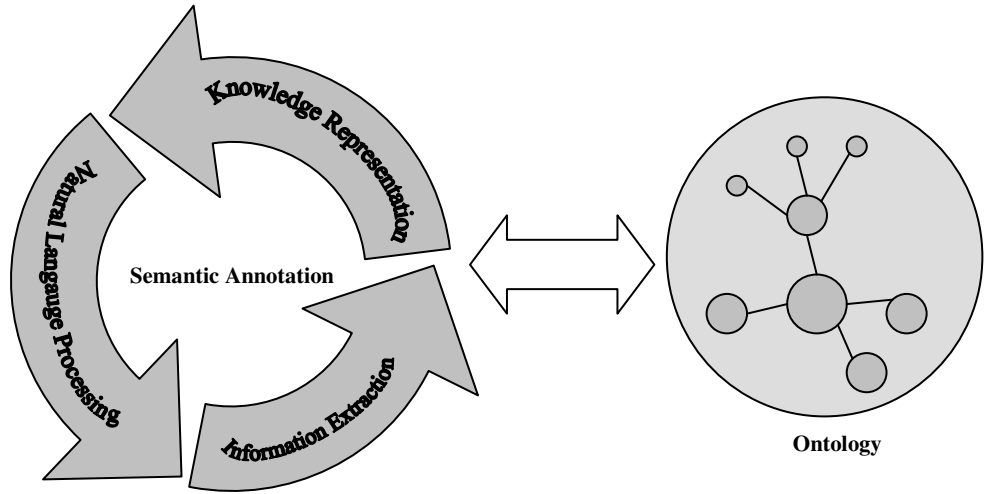

Figure 1. An overview of semantic annotation and effective technologies

## 3. RELATED WORKS

Different semantic annotation tools and systems have been developed during years after advent of semantic web technology. These tools and systems which are called semantic annotation platforms can be classified based on the type of annotation method used in them [12]. There are two primary categories, Pattern-based and Machine Learning-based, as shown in Figure 2. In addition, platforms can use methods from both types of categories, called Multistrategy, in order to take advantage of the strengths, and compensate for the weaknesses, of the methods in each category.

Pattern-based approaches can perform pattern discovery or have patterns manually defined. Most of these methods follow the process in which an initial set of entities is defined at the beginning and the corpus is scanned to find the patterns that contain the entities. New entities are discovered, along with new patterns. This process continues recursively until no more entities are discovered or the user stops the process. Annotation can also be generated by using manual rules to find entities in text [12].

Machine learning-based semantic annotation platforms utilize two methods: probability and induction. Probabilistic semantic annotation platforms use statistical models to predict the locations of entities within text.

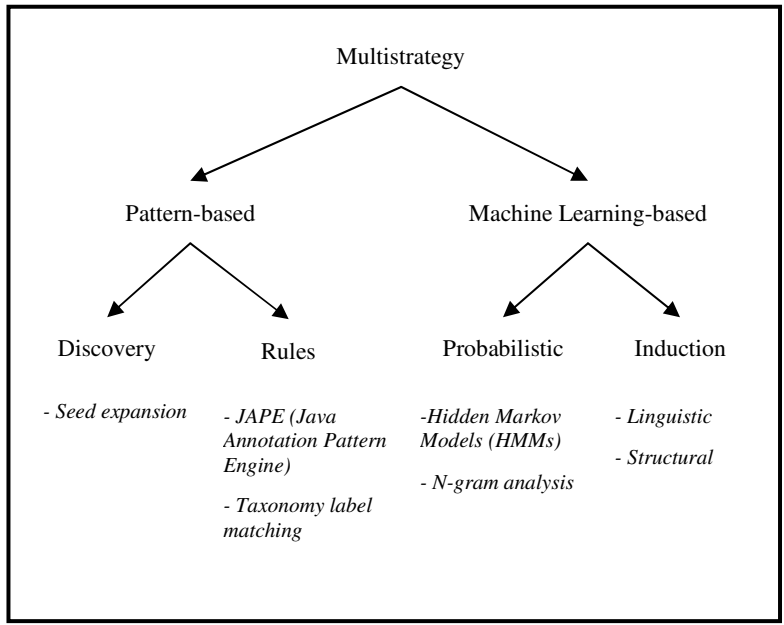

Figure 2. An overview of semantic annotation and effective technologies [12]

## 4. CLASSIFICATION OF SEMANTIC ANNOTATION CHALLENGES

There are many challenges and obstacles in semantic annotation domain that lead to several research opportunities. The process of semantically annotating documents is a well-known challenge for the Semantic Web per se [14]. So, in this section we present an analytical review of all these obstacles. Regarding to different problems that emerge for enhancing the content of current web pages and developing domain ontoligies, we classify the challenges of semantic annotation systems into two inclusive classes: a) general challenges, and b) technical challenges. The general challenges category refers to those obstacles that exist regardless of technical and algorithmic considerations, such as multilinguality and scalability problems. But technical challenges contain the problems that relate to implementation and performance of a semantic annotation system [5,12,15-19]. Figure 3 reveals a comprehensive classification of semantic annotation challenges.

The general challenges category is divided into two classes of linguistic and content related obstacles. Multilinguality means that the contents of web pages are written in different languages. This characteristic is a hurdle against making a general and comprehensive annotation system. In addition, whereby ontologis are created in different languages, this makes some problem for annotation and communication between ontologies. Standardization of semantic annotation languages is another challenge in this category. By standardizing these languages it would be possible to reach a consistency and homogeneity among web pages. With standardizing the output structure of semantic annotation systems, it would be able to face the problem of multilinguality by using automatic translation tools that work on these standard structures.

In the content related challenges category, the semantic annotation obstacles are divided into three classes; heterogeneity of documents format, scalability, and dynamic documents. Diversity in web pages is common feature in web domain, but in semantic annotation area, this characteristic turn into a drawback. This feature can make problems for creating semantic annotation tools even in a specific domain.

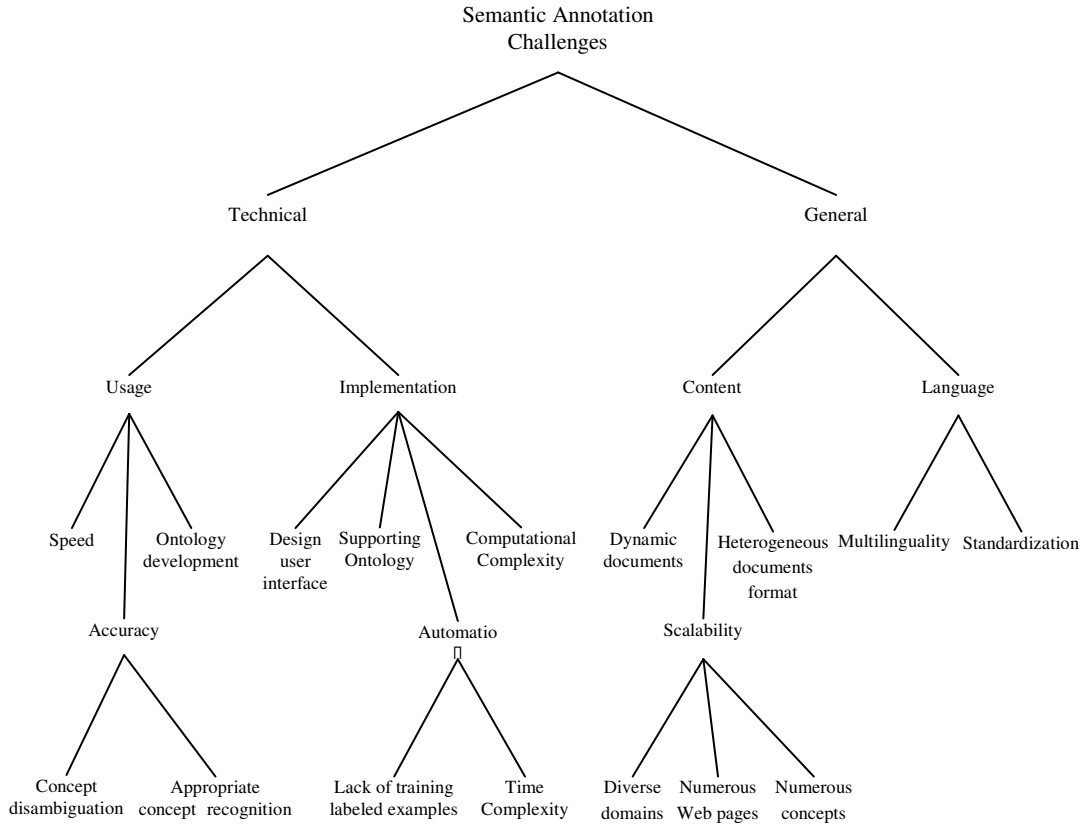

Figure 3. Classification of Semantic Annotation Challenges

One of the other web characteristic that has the same impact like heterogeneity on semantic annotation is dynamic feature of web pages, that is, the continuous changing and updating of web pages. Handling large volume of web documents is another challenge that semantic annotation systems must deal with it. Another factor that intensifies this problem is the wide range of domains in the content of the web pages. While the goal of semantic annotation process is recognizing concepts and adds the proper semantic metadata to the web pages, it's obvious that handling different domains is a serious problem in this field.

The technical challenges are categorized into two main classes; implementation challenges and usage challenges. The first and the most important challenge in the field of semantic annotation is automation of annotation systems. As we mentioned before, manual annotation is an expensive and time consuming task. Also, to annotate large amount of web documents with different domains, it is crucial to automate the process of semantic annotation. Another obstacle in this class of challenges is computational complexity of annotation algorithms especially NLP based approaches. While semantic annotation systems want to face a large scale problem, they must have an acceptable time performance.

Other class of challenges is emerged after the development of annotation systems. In our classification we categorized these challenges under the group of usage challenges. Due to large amount of resources that semantic annotation systems face with them, they must perform the annotation process in an acceptable time. On the other hand, there may be many ambiguous concepts that annotation systems must recognize them correctly, so disambiguation of concepts is another problem that must be handling by these systems. During annotation process it is possible to extract new concepts that there would be no definition for them in the domain ontology. It's an effective feature for an annotation system to be able to add these concepts to the ontology.

## 5. ANALYTICAL FRAMEWORK

In this section we present the analytical framework show the efficiency of different machine learning applications in addressing some of the semantic annotation challenges. This framework is introduced in Table 1. It tries to reveal the relation of effective machine learning techniques to deal with some semantic annotation challenges and development of the annotation systems. The classification of semantic annotation systems in this framework is based on machine learning approaches, so there are five classes:

- Supervised Learning
- Unsupervised Learning
- Semi-supervised Learning
- Active Learning
- Hybrid of semi-supervised learning and active learning

In an extensive and fast changing research area such as semantic annotation, which itself is a cornerstone for semantic web, it is not possible to review all the approaches and tools that continuously are introduced every day. So in this paper we only select some approaches that are related to different machine learning techniques. For this goal, we try to mention some well-known and new methods in this area.

In our framework some of annotation systems are reviewed; a concise description of systems is shown and the way that these systems utilize ontologies is described. In addition, we study these systems regard to the way that their annotation process are performed, i.e. manual, semi-automatic, or automatic. And then, the scope of these systems is described briefly.

Table 1. Analytical Framework of Machine Learning Approaches for Semantic Annotation

| | Semantic Annotation Systems | Description | Ontology Development | Automation | Application Scope |
|---|---|---|---|---|---|
| **Supervised Learning** | Action [19] | Uses classification for determining and separating different events | Supporting Ontology | Manual | Domain dependent |
| | RCSSAT [20] | Relation classification by using a new lexicon to provide semantic behavior features of words, and using kernel method to model lexical features | - | Manual | General |
| | AnnoTex [21] | Annotating based on classifying documents by means of semantic similarities | Supporting Ontology | Manual | Domain dependent |
| | KZMCM [22] | Using text mining for semi-automatically semantic annotation | Supporting Ontology | Semi-automatic | Domain dependent |
| | SOZEKAMM [23] | Automation of the generation of an annotation schema for a given semantic domain using a supervised categorical clustering algorithm LIMBO | - | Semi-automatic | Domain dependent |
| | CAFETIERE [24] | Using text mining techniques to propose annotation suggestions | Supporting Ontology | Semi-automatic | Domain dependent |
| **Unsupervised** | BroMo [25] | Using clustering for blogs and article semantic annotation | - | Semi-automatic | General |
| | OEAKM [26] | Built an ontology enabled annotation and knowledge management system that provides clustering and real-time discussion for collaborative learning | Supporting Ontology | Semi-automatic | General |
| | PARMENIDES [27] | Using clustering for the establishment of ontologies and the semantic annotation of documents with the concepts, entities and events depicted in the ontologies | Supporting and enhancing Ontology | Semi-automatic | General |
| | ASWSACC [28] | A machine learning-based semantic web annotation tool that learns by mining association rules among words through the text. | Supporting Ontology | Semi-automatic | General |
| | EOAAC [29] | Using association rule mining to extract co-occurrences of concepts | Supporting and enhancing Ontology | Semi-automatic | Domain dependent |
| **Semi-supervised Learning** | Self-teaching SVM-struct [30] | Proposing a novel self-teaching SVM-struct model to improve the performance of semantic annotation with fewer labeled examples | Supporting Ontology | Automatic | General |
| | LVNER [31] | Presenting a simple semi-supervised learning | - | Automatic | General |
| | | algorithm for named entity recognition using conditional random fields (CRFs) | | | |
| **Active Learning** | ASCUM [32] | Proposing a SVM-struct based active learning algorithm for automatic semantic annotation | Supporting Ontology | Semi-automatic | General |
| **Hybrid of Semi-supervised and Active Learning** | TM [33] | A hybrid approach that annotate confident samples automatically and leave other uncertain samples to be labeled by a human annotator | - | Semi-automatic | General |
| | LSWW [34] | Proposing a combination of active learning and self-training method to reduce the labeling effort for Chinese Named Entity Recognition and Annotation | - | Semi-automatic | General |
| | 1L-SP SSAL [35] | A token level combination of semi-supervised and active learning with a variance based confidence measure | - | Semi-automatic | General |

# 6. CONCLUSION

In this paper, we present an inclusive layered classification of semantic annotation challenges. This classification represents almost all of the challenges that are mentioned in various researches. Due to the wide range of domains and the dynamic environments that the semantic annotation systems must be performed on, we discussed that automating the annotation process is a vital requirement for semantic annotation systems. So, automation is one of the most serious challenges in this field. Then we reviewed and analyzed machine learning applications for solving semantic annotation challenges such as ontology development, scalability, and more specifically the automation problem.

From this point of view, we presented an analytical framework regarding these applications. In this framework some of the annotation systems based on the important features in this domain are reviewed. Results show that different learning approaches have great impact to solve semantic annotation challenges. Whereby most of the systems use supervised and unsupervised techniques in their methods, it seems that more researches are required to be directed to the applications of newer learning techniques such as semi-supervised and active learning. Also, preparing labeled corpora for training learner's models is one of the significant issues in many text based tasks, so approaches such as semi-supervised learning and active learning that deal with reduction of labeling costs can be very efficient in semantic annotation systems. Furthermore, it's shown that a combination of these two approaches can outperform many individual systems.

**Authors**

Hamed Hassanzadeh received his B.S. in Software Engineering from Islamic Azad University, Lahijan Branch, Lahijan, Iran. Currently, he is pursuing M.S. in Software Engineering at Islamic Azad University, Qazvin Branch, Qazvin, Iran. His research interests include Semantic Web and Machine Learning.

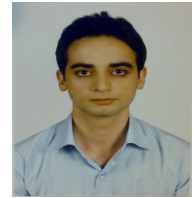

MohammadReza Keyvanpour is an Assistant Professor at Alzahra University, Tehran, Iran. He received his B.S. in Software Engineering from Iran University of Science &Technology, Tehran, Iran. He received his M.S. and Ph.D. in Software Engineering from Tarbiat Modares University, Tehran, Iran. His research interests include image retrieval and data mining.

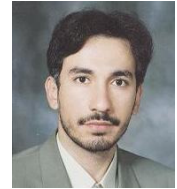